# Semi-Supervised Graph Embedding for Multi-Label Graph Node Classification

Kaisheng Gao, Jing Zhang, Cangqi Zhou

Nanjing University of Science and Tenchnology, China
`kaisheng_gao@njust.edu.cn`

**Abstract.** The graph convolution network (GCN) is a widely-used facility to realize graph-based semi-supervised learning, which usually integrates node features and graph topologic information to build learning models. However, as for multi-label learning tasks, the supervision part of GCN simply minimizes the cross-entropy loss between the last layer outputs and the ground-truth label distribution, which tends to lose some useful information such as label correlations, so that prevents from obtaining high performance. In this paper, we propose a novel GCN-based semi-supervised learning approach for multi-label classification, namely ML-GCN. ML-GCN first uses a GCN to embed the node features and graph topologic information. Then, it randomly generates a label matrix, where each row (i.e., label vector) represents a kind of labels. The dimension of the label vector is the same as that of the node vector before the last convolution operation of GCN. That is, all labels and nodes are embedded in a uniform vector space. Finally, during the ML-GCN model training, label vectors and node vectors are concatenated to serve as the inputs of the relaxed skip-gram model to detect the node-label correlation as well as the label-label correlation. Experimental results on several graph classification datasets show that the proposed ML-GCN outperforms four state-of-the-art methods.

## 1 Introduction

There exist many graph-structured datasets in the real world, such as social networks, academic citation networks, and knowledge graph. Graph Representation Learning (GRL) methods that aim to learn the vector representations for graphs has attracted much attention in recent years. Because the dimension of every node vector could be very large it may suffer from the high computational complexity and huge memory space usage, if we merely use the 1-hot encoding methods or a discrete adjacency matrix to present the nodes. Therefore, we usually embed a graph into a low-dimensional space, which not only preserves the structural information but also significantly reduces the computational costs. Within this low-dimensional space, the distance between two nodes with the close relation in the original graph will also be close in a measure derived from the embedding presentation. Here, the close relation of two nodes means that they are directly connected with each other or share a set of common neighbors, which is often used to define the similarity of two users in a social network.

There are several graph embedding methods proposed in recent years. For example, GF [1] factorizes the adjacency matrix and minimizes the L2-norm of the embedding matrix. LINE [2] defines two joint probability distributions for each pair of nodes, one using the adjacency matrix and the other using the embedding vector. Then, LINE minimizes the KL divergence of these two distributions. DeepWalk [3] uses random walk to generate a node sequence. Then, for each node sequence, it applies the Word2Vec model [4] to get the node embedding by treating each sequence as a word sentence. All the above methods can be classified as the shallow model, compared with the methods using deep learning technology. Recently, a kind of deep learning model, namely graph neural network (GNN), has attracted much attention, including some typical methods of GraphSage [5], GAT [6] and GCN [7], which use neural networks to train classification models on graph-structure datasets.

Graph convolutional network (GCN) is a deep neural network model to catch structural information in a graph, which has been widely used in several machine learning paradigms, such as semantic role labeling [8], event extraction [9] and recommendation task [10]. In additional, the GCN model also obtain good performance in graph-based semi-supervised learning because its structure is robust to the missing information in training sets [7]. In a semi-supervised learning task, GCN uses a graph convolution operation to integrate each node and its one-hop neighbor information in each layer. After conducting several layers of convolution, each node can gather its $k$-hop neighbor information in the final layer, which is the embedded feature presentation of such a node. Then, we can use some supervised information to train a classifier based on these embedded features.

Usually, multi-label classification models are trained in a semi-supervised manner, because not all labels on every instance are obtained values. In multi-label graph datasets, one node may have several labels. i.e. the correlation between this node and these labels are high, we called it node-label correlation. if two labels are highly correlated, the nodes with these labels should be close in the embedding space. For example, in movie genres dataset, the genres (labels) 'Western' and 'Adventure' always appear in a same movie. Thus, two movies with labels 'Western' and 'Adventure' respectively should also be close. We called this *label-label correlation*. Because this correlation is not reflected in the graph structure it cannot be captured in the original GCN models. Accordingly, for a multi-label graph data, some nodes may have several specific labels. That is, one node and some labels may be highly correlated, which is called the *node-label correlation* in this study.

To address this issue, we propose a novel GCN-based model for semi-supervised multi-label graph node classification, namely ML-GCN. To capture the high non-linear correlations among nodes, we use a two-layer neural network model, on each of which we conduct a series of graph convolution operations. To preserve the label-label correlation, we treat each label as a vector so that we can measure the relationship between two labels. After labels are embedded, we can shorten the distance be-tween two nodes whose labels are highly correlated in the embedding space. After obtaining the representation of all nodes in a graph, we can train a multi-label classifier to make predictions to the unlabeled nodes. In the proposed ML-GCN method, we use a sigmoid layer as the downstream learning method. The contributions of this paper are three-fold:

- We first investigate the applicability of graph convolutional network applying to the multi-label learning and point out that the label-label correlation should be considered to improve the learning performance.
- We propose a novel learning method ML-GCN, where labels on nodes and the nodes themselves are uniformly embedded into a same low-dimensional space. ML-GCN can capture both node-label and label-label correlations. To the best of our knowledge, it is the first that the labels of each node are embedded and fed into GCN.
- We conduct a comprehensive empirical study on three real-world multi-label graph node classification datasets, whose results demonstrate that ML-GCN outperforms four state-of-the-art methods.

The remainder of the paper is organized as follows: In Section 2, we briefly review the related work. Section 3 presents the novel ML-GCN method. In Section 4, we compare our ML-GCN with four state-of-the-art methods on several real-world datasets. Section 5 concludes the paper with some future work.

## 2    Related Work

A large number of application problems can be abstracted into the classification problem of nodes in a graph structure. In recent years, various kinds of graph neural network models have been proposed [11], including Graph Convolution Networks [7], Graph Attention Networks [12], Graph Autoencoder [13], Graph Generative Networks [14], Graph Spatial-temporal Networks [15] and so on. The principle of most of these approaches is "neural message passing" proposed by Gilmer et al. [16]. In the message passing framework, a GNN can be viewed as a message passing algorithm, where the representation of a node is iteratively computed from the features of its neighbor nodes using a differentiable aggregation function. For the identity of principle, the GCN model can be considered as the fundamental of most GNN models [11] which aggregates each node with its neighbors and let the node receive messages from its neighbors. Therefore, in this paper, we mainly focus on the GCN model. GCN can be divided into two categories: spectral-based and spatial-based approaches. The spectral-based methods define convolution operations by introducing filters from the perspective of graph signal processing [17], where the convolution on the graph is interpreted as removing noise from graph signals and passing message in the spectral domain. The spatial-based approaches formulate convolution operations on a node as aggregating feature derived from its neighbors and the information passing through it. In general, all the GNN-based methods attempt to embed the graph structural information into vectors and follow the same hypothesis that nodes with similar structure tend to be close in the embedding space.

Multi-label learning is usually semi-supervised because, in many situations, instances in the training set do not necessarily have all the potential labels been assigned values. The training process usually learns from fully-labeled, partly-labeled, and even unlabeled samples to form predictive models. For the multi-label learning in a graph structure, a straightforward method is to train multiple independent binary classifiers for each label. However, this simple method has several defects: It does not consider the correlations among labels; The number of labels to predict will grow exponentially as the number of label categories increases; It is essentially limited by ignoring the topological structure among nodes. In some recent studies, researchers attempted to capture label-label correlations in some classical deep learning models for multi-label classification. Gong et al. [23] used a ranking-based learning strategy to train deep convolutional neural networks for multi-label image recognition and found that the weighted approximated-ranking loss performs best. Wang et al. [24] utilized recurrent neural networks (RNNs) to transform labels into embedded label vectors, so that the correlation between labels can be employed. Wang et al. [25] introduced a spatial transformer layer and long short-term memory (LSTM) units to capture label correlation.

In this study, our novel learning method is still based on the GCN model but first introduces the label matrix embedding to capture the label-label correlation among the graph nodes.

## 3 The Proposed Method

The key idea behind the proposed ML-GCN is that it embeds multiple labels and nodes in the same space, where label-label correlations and label-node correlations can be simultaneously considered. In this section, we first introduce the problem statement and some preliminaries. Then, we present the label embedding scheme of ML-GCN. Finally, we present the optimization algorithm of the ML-GCN model.

### 3.1 Problem Statement

We define an $G = (V, E, X, Y)$ as an undirected graph, where $V = (V_l \cup V_u)$ is a finite node set that includes $n_l$ labeled nodes ($V_l$) and $n_u$ unlabeled nodes ($V_u$). There are $n = n_l + n_u$ nodes in total. $E$ is an edge set and $X \in \mathbb{R}^{n \times d}$ is a feature matrix of all the graph nodes. $Y \in \mathbb{R}^{n_l \times c}$ is a 0-1 matrix that presents the labels of $n_l$ labeled nodes, where $c$ is the maximum number of labels on each instance. The adjacency matrix of the graph is denoted by $A = [a_{ij}] \in \mathbb{R}^{n \times n}$, where $a_{ij}$ is the weight assigned on the edge between nodes $i$ and $j$. The degree matrix of $A$ is denoted by a diagonal matrix $D = diag(d_1, \ldots, d_n)$, where $d_i = \sum_j a_{ij}$ is the degree of node $i$. The symmetric normalized Laplacian matrix is denoted by $L_{sym} = I - D^{-\frac{1}{2}} A D^{-\frac{1}{2}}$. Our goal is to build a multi-label classification model that can predict the labels of unlabeled graph nodes.

### 3.2 Preliminaries: Graph Convolutional Network

To embed features of nodes and their structural information, we first introduce a graph convolutional network [7]. In particular, the core of GCN is the operation in each layer, which can be defined as:

$$H^{(l+1)} = \sigma\left(\widetilde{D}^{-\frac{1}{2}}\widetilde{A}\widetilde{D}^{-\frac{1}{2}}H^{(l)}W^{(l)}\right). \tag{1}$$

Here, $\widetilde{A} = A + I_{n_l+n_u}$ is an adjacency matrix with self-connections added. Matrix $I_{n_l+n_u}$ is an identity matrix. Diagonal matrix $\widetilde{D} = diag(\tilde{d}_1, ... \tilde{d}_n)$ is a degree matrix of $\widetilde{A}$, where $\tilde{d}_i = \sum_j \widetilde{A}_{ij}$. $W^{(l)}$ the trainable parameters of the $l$-th layer. Function σ is an activation function. In this paper, the activation function of each layer is defined as $\sigma(x) = \max(0, x)$ as it used in other studies [7]. In the first layer, we have $H^{(0)} = X$. That is, we take the graph feature matrix as the input of GCN. In the last layer, we have

$$Y_{prob} = \text{sigmoid}(H^{(l+1)}), \tag{2}$$

Where $H^{(l+1)} \in \mathbb{R}^{n \times c}$ and $Y_{prob}$ is the probability distribution of labels for each node. Then, we minimize the cross-entropy loss between $Y_{prob}$ and labeled nodes:

$$\min \sum_{i=1,2,...,n_l} y_i log(y_{prob}^{(i)}). \tag{3}$$

where $y_i$ and $y_{prob}$ denote the row vectors of Y and $Y_{prob}$. That is, we embed all nodes into a $c$-dimension space and use a sigmoid function to determine the prediction of the labels. However, this simply model may be confronted with some drawbacks:

- If we utilize fewer layers to construct GCN, the difference between the dimension of the last layer and the second to the last layer may be quite large. It may cause the hidden feature loss and make the model difficult to optimize. For example, on the *Citeseer* dataset [26] whose input feature dimension is 3703 and the number of labels is 6, if we use a two-layer GCN, we cannot let the dimension decrease smoothly regardless of the settings of the hidden layer dimension.
- As [18] pointed out, if we simply stack more layers, the model will mix the features of nodes from different labels and make them indistinguishable. This is because each layer of GCN applies Laplacian smoothing [19] to features, and every two nodes with a connected path tend to be close with Laplacian smooth.
- A multi-label classification model with a sigmoid layer cannot capture the label-label relationship because it treats each label individually. Thus, it may lose some information on the multi-label graph dataset.

### 3.3 ML-GCN: Label Embedding Matrix

The proposed ML-GCN introduces a label embedding matrix as well as the label-node co-embedding to GCN. Let $Z_Y \in \mathbb{R}^{n_l \times l}$ denote the label embedding matrix, where $n_l$ is the number of labeled nodes and $l$ is the dimension of label vectors. We generate label embedding matrix randomly at the beginning of training. The dimension of the matrix is the same as the dimension of node features before the last graph convolution operation. Here, we set $H^{(l+1)}$ as the last output before the sigmoid layer. That is, the dimension $l$ of label embedding matrix is the same as the dimension of $H^{(l)}$. Then, we can calculate the label-label correlation and the label-node correlation using the $Z_Y$ and $H^{(l)}$, respectively. Figure 1 shows the framework of the proposed

ML-GCN. Here, each grid represents a matrix. We feed a graph into the first GCN layer and obtain the first embedding matrix as the output of this layer. Then, we use the randomly generated label embedding matrix to calculate the label-label loss, and together with the first embedding matrix to calculate the label-node loss. Then, we feed the first embedding matrix into the second GCN layer. Finally, we use the output of the sigmoid layer to calculate the cross-entropy loss against the ground truth.

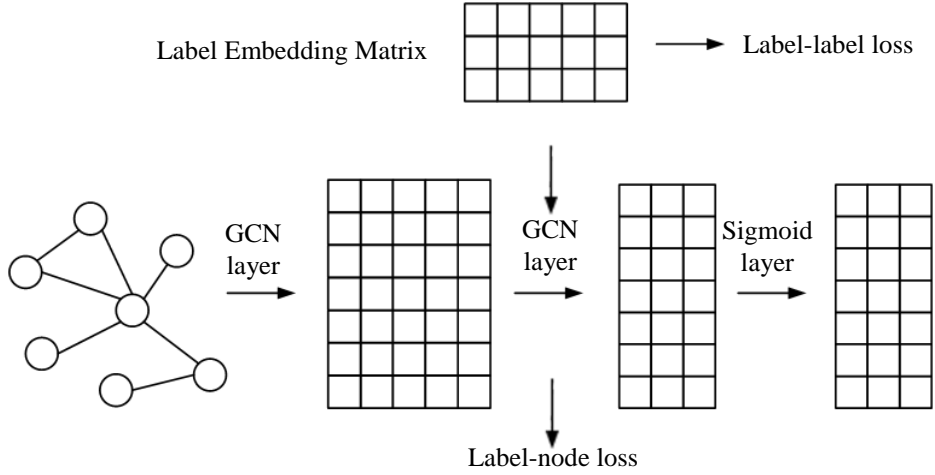

**Fig. 1.** The framework of the proposed ML-GCN.

Consider a node with several labels. Our goal is to maximize the occurrence probability of these labels given the node. The inputs are the node vectors and the corresponding label vectors. If we treat a node and its labels as a sentence, our goal also can be expressed as "given a center word (node), to predict the neighbor words (labels)," which is the essential idea of Skip-gram[20]. For example, in Figure 2, we have a node with four labels, and we can treat each element as a word and generate a sentence. Then, we utilize the Skip-gram for the next calculate.

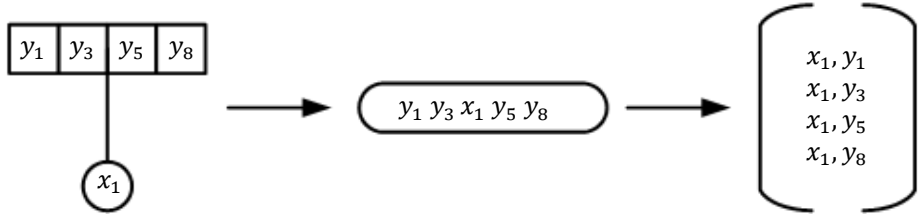

**Fig. 2.** Convert a node with several labels to a sentence

In the Skip-gram model, for a word $w_i$ and window size $c$, we can extract $w_i$ and its $c$-$1$ neighbors with $w_i$ at the center. Word $w_i$ and each of its neighbor can form a pair as $(w_i, w_j)$. The co-occurrence probability of $w_j$ given $w_i$ is defined as:

7$$P(w_j|w_i) = \frac{\exp(w_j^T w_i)}{\sum_{t=1}^{M} \exp(w_t w_i)}, \quad (4)$$

where $M$ is all the words in the corpus. Thus, we can obtain the word embedding by maximize such co-occurrence probability for all the word pairs.

Consider the node-label sentences. Given a node $x_i$ and its labels $Y_{x_i} = \{y_1, y_2, \ldots, y_c\}$, the vector representation of $x_i$ is the $i$-th row of $H^{(l)}$, denoted by $h_i$. The label vector of $y_j$ is $z_{y_j}$. We only consider the node as the center word and remove the window size. We use each label to form a pair with the node because there is no predefined order of its labels. Therefore, we have a set of node-label pairs, denoted by $\{(x_iy_1), (x_iy_2), \ldots, (x_iy_c)\}$. For any node $x_i$, we can optimize the node and its label embedding by maximize the object function as follows:

$$\max \frac{1}{c} \sum_{y_j \in Y_{x_i}} \log P(z_{y_j}|h_i). \quad (5)$$

Since this function is operated in the second to the last layer of GCN and uses the features of layer $H^{(l)}$, we can better capture the node-label correlation in a high dimensional space before the feature dimension is reduced to the label-class wise. As we know GCN conducts the Laplacian smoothing on each node, whose consequence is that the presentations of many nodes may tend to be the same at the final stage of training. Adding this function prevents the side-effort of the Laplacian smoothing in GCN. It hinders the Laplacian smoothing which aggregate each node to be hard to distinguish. Thus, it can accelerate the training process and prevent the model from over smoothing that makes each node converge to the same point.

To capture the label-label correlation, we utilize the same model but get rid of the node vectors. That is, we only use the labels of a node to construct the sentence. For example, given a node $x_i$ with labels $\{y_1, y_2, \ldots, y_c\}$, we only use labels to construct a sentence, which forms a set of label-label pairs, denoted by $\{(y_1y_2), (y_1y_3), \ldots, (y_cy_{c-1})\}$. Note that the pairs of $(y_iy_j)$ and $(y_jy_i)$ are different. similar to Eq. (5), we have the object function:

$$\max \frac{1}{c} \sum_{y_i, y_j \in Y_{x_i}, i \neq j} \log P(z_{y_j}|z_{y_i}). \quad (6)$$

If the node only has one label, we omit the label-label relation and only calculate Eq. (5) on this node. To maximize the Eq. (5) and Eq. (6), we can reserve the node-label correlation as well as label-label correlation in the embedding space.

### 3.4 CO-Optimization and Negative Sampling

To calculate Eq. (5) and Eq. (6), we need to calculate $P(z_{y_j}|h_i)$ and $P(z_{y_j}|z_{y_i})$, which requires the summation over all the labels. The calculation may cost too much running time because some multi-label graph datasets may have abundant label types. To accelerate the calculation of these two co-occurrence probabilities, we use a trick of negative sampling in the Skip-gram model. First, we rewrite Eq. (1) as follow:

$$\min -\log \sigma \left(z_{y_j} h_i\right) - \sum_{t=1}^{K} \mathbb{E}_{y_t \sim P(y)} \log \sigma \left(-z_{y_t} h_i\right), \quad (7)$$

where $K$ is a hyper parameter denoting the number of sampled labels for one node-label pair. Thus, the task becomes to distinguish the target label $y_j$ from $K$ labels drawn from the noise distribution $P(y)$. The idea behind negative sampling is: We want to maximize the co-occurrence probability of $z_{y_j}$ given $h_i$ and minimize the probability of a randomly sampled labels $z_{y_t}$ given the same node $h_i$. In practice, we define a noise distribution as chosen to be $U(y)^{3/4}/\sum_y U(y)^{3/4}$, where $U(y)$ is the unigram distribution of the labels. Here, we only consider the co-occurrence times of each label type on labeled data as the unigram distribution. If the sample process obtains the positive label $y_t = y_j$, we just resample $y_t$ until the condition $y_t \neq y_j$ is satisfied.

Similar to Eq. (5), we sample $K$ labels as the negative labels and rewrite the Eq. (6) as follows:

$$\min -\log \sigma \left(z_{y_j} z_{y_i}\right) - \sum_{t=1}^{K} \mathbb{E}_{y_t \sim P(y)} \log \sigma \left(-z_{y_t} z_{y_i}\right), \quad (8)$$

The goal is to distinguish the label $y_j$ from $K$ sampled negative labels on the condition of given $y_i$. To calculate Eq. (7) and Eq. (8) in each labeled node, we can obtain the loss function $L_{n-l}$ denoting the node-label loss calculated by Eq. (7) and $l_{l-l}$ denoting the label-label loss calculated by Eq. (8). With the sigmoid loss of the last layer, we can have the final objective for optimization:

$$L_{sum} = \lambda_1 L_{l-l} + \lambda_2 l_{n-l} + l_{sigmoid}, \quad (9)$$

where $\lambda_1, \lambda_2 \in \mathbb{R}$ are the hyper parameters to weight three terms in the objective functions. We optimize the function with Adam optimizer [28]. We summarize all above contents with a pseudocode and list as follow:

---

**Algorithm 1**: ML-GCN (Training and Predicting)

**Input**: Graph $G$, feature $X$, label $Y_L$, number of GCN layers $l+1$
**Output**: labels of unlabeled nodes $Y_U$
1: randomly generate the label matrix $Z_Y$
2: $H^{(0)} = X$
3: **for** epoch = 1, …, n **do**:
3:    **for** i = 0, 1, …, l:
4:        calculate the output of $i+1$ GCN layer $H^{(i+1)}$ using $H^{(i)}$
5:    $L_{sigmoid} = \text{crossentropy}\left(Y_L, \text{sigmoid}(H^{(l+1)})\right)$
6:    calculate Eq. (7) using $Z_Y$ and $H^{(l)}$, obtain $L_{n-l}$
7:    calculate Eq. (8) using $Z_Y$, obtain $L_{l-l}$
8:    optimize to minimize $L_{sum} = \lambda_1 L_{l-l} + \lambda_2 l_{n-l} + l_{sigmoid}$
9: **Return** 1 if $Y_{prob} = \text{sigmoid}(H^{(l+1)})$ is greater than 0.5, otherwise 0, given $Y_U$

## 4 Experiments

In this section, we first present the datasets used in our experiments, methods in comparisons, and the experimental settings. Then, we focus on discussing the experimental results.

### 4.1 Datasets

Compared with the plenty of single-label classification datasets, there are only a few real-world multi-label graph node classification datasets that can be used in our experiments. We evaluate our ML-GCN model on three datasets collected from different domains of biology, movie, and social media. These datasets are chosen not only because they belong to different domains but also, they have different network topologic structures. The details of the datasets are listed in Table 1.

**Table 1.** The details of the datasets used in our experiments

| Dataset | Domain | Nodes | Edges | Classes | Features |
|---------|--------|-------|-------|---------|----------|
| *Facebook* | Social | 347 | 5038 | 24 | 224 |
| *Yeast* | Biology | 1240 | 1674 | 13 | 831 |
| *Movie* | Movie | 7155 | 404241 | 20 | 5297 |

The *Facebook* dataset [28] is a social network. The nodes represent users of Facebook and the edges represent the fan following relation. The feature of each node is the personal information of the corresponding user. The task is to determine the 'circles' tags of each user (node). One user can belong to multiple circles.

The *Yeast* dataset is part of the KDD Cup 2001 challenge [29]. The graph is constructed based on the interactions between proteins. Each node represents a gene. The gene code information is set as the feature of nodes. The task is to predict the function of these genes.

We constructed a movie dataset from Movielens-2k dataset [30]. The Movielens-2k dataset contains movies information such as actors, genres, and tags information. We set the tags information as the feature of movies and set a common director as an edge. For example, if two movies share the same director, we added an edge between these two movies, and set the weight of this edge 1. The task is to predict the genres of the movies.

### 4.2 Methods in Comparison and Experimental Settings

**Method in Comparison:** We compared our ML-GCN with the following state-of-the-art methods:
- Multilayer perception (MLP) is a classical label classifier takes only node feature as input and ignores the graph structure.
- Deepwalk [3] learns node features by treating random walks in a graph as the equivalent of sentences.
- GCN [7] takes both node feature and graph structure as the input.

- Partly ML-GCN is a simpler ML-GCN without the calculation of the label-label loss. This method is added to evaluate the impact of the loss function on the performance of the learning models.

**Experimental settings:** For fair comparisons, all the methods (MLP, GCN, Partly ML-GCN and ML-GCN) use two-layer models. For dataset *Facebook*, we set the middle layer dimension to 64 and use 100 nodes for training and 150 nodes for testing. For dataset *Yeast*, we set the middle layer dimension to 256 and use 200 nodes as training nodes and 500 nodes as testing nodes. For dataset Movie, we set the middle layer dimension to 512 and use 500 nodes as training data and 2000 nodes as testing nodes. For all datasets, we set the number of negative sample to 5, set walk length to 40 for DeepWalk and set the window size to 10. All the models are trained using Adam [29] with a learning rate of 0.01. The parameters $\lambda_1$ and $\lambda_2$ are both equal to 0.25. We use the micro-F1 score (in percentage) as the evaluation metric in the paper.

### 4.3 Experimental Results

**Experiment 1 (Overall Performance):** The classification results of five methods on three datasets in terms of the micro-F1 score are summarized in Table 2. We have the following observations. Overall, our proposed ML-GCN method consistently outperforms the other methods on all datasets. Compared with the original GCN, on dataset *Facebook*, our ML-GCN achieves the improvement of 1.72 points, and on datasets with stronger label-label correlations (i.e., datasets *Yeast* and *Movie*), the improvement of ML-GCN archived as high as 3 points. Thus, ML-GCN successfully captures the label-label correlations. Furthermore, our ML-GCN also outperforms the Partly ML-GCN on all dataset, which shows that the calculation of the label-label loss in the model training indeed improves the performance of the learning models.

**Table 2.** Experimental results in terms of micro-F1 score (in percentage)

| Method | *Facebook* | *Yeast* | *Movie* |
|---|---|---|---|
| MLP | 58.13 | 63.79 | 33.62 |
| DeepWalk [3] | 58.89 | 53.40 | 33.94 |
| GCN [7] | 58.13 | 63.16 | 35.72 |
| Partly ML-GCN | 59.51 | 65.27 | 37.75 |
| ML-GCN | **59.85** | **66.06** | **37.96** |

**Experiment 2 (Performance under Different Training Set Sizes):** To investigate whether our ML-CGN is consistently superior to GCN under different training set sizes, we randomly selected different proportions of the instances from the original datasets to form the training sets. The experimental results are summarized in Table 3. We have the following observations.

**Table 3.** Experimental results under different training set sizes in terms of micro-F1 score (in percentage)

| Dataset | Method | 10% | 20% | 30% | 40% |
|---|---|---|---|---|---|
| *Facebook* | GCN [7] | 57.25 | 58.45 | 59.95 | 60.05 |
| | ML-GCN | 58.13 | 59.63 | 60.14 | 60.98 |
| *Yeast* | GCN [7] | 61.23 | 62.45 | 62.73 | 63.68 |
| | ML-GCN | 63.03 | 64.54 | 64.04 | 65.77 |
| *Movie* | GCN [7] | 36.82 | 37.64 | 38.06 | 38.23 |
| | ML-GCN | 38.04 | 39.92 | 40.64 | 40.76 |

Overall, the proposed ML-GCN outperforms GCN under all differ-ent training set sizes on all datasets. On the Movie dataset, the advantage of ML-GCN over GCN will increase as the proportion of the training sets increases. That means, when the training in-stances increase, our ML-GCN is easier to capture the label-label correlations.

## 5    Conclusion

In this paper, we present a novel ML-GCN method for semi-supervised multi-label graph node classification. By embedding the label and node information into the same low-dimensional space, ML-GCN can jointly capture both node-label and label-label correlations, which improves the performance of the learning models, compared with the state-of-the-art methods. In the future, we will consider embedding the contents of nodes to the learning models.